\documentclass{sig-alternate-05-2015}

\usepackage{amsmath}
\usepackage{multirow}
\usepackage{blkarray}
\usepackage{graphicx}
\usepackage{subfigure}
\usepackage{adjustbox}

\begin{document}

% Copyright
% \setcopyright{acmcopyright}
%\setcopyright{acmlicensed}
%\setcopyright{rightsretained}
%\setcopyright{usgov}
%\setcopyright{usgovmixed}
%\setcopyright{cagov}
%\setcopyright{cagovmixed}

\CopyrightYear{2016}
\setcopyright{acmcopyright}
\conferenceinfo{UCC '16,}{December 06-09, 2016, Shanghai, China}
\isbn{978-1-4503-4616-0/16/12}\acmPrice{\$15.00}
\doi{http://dx.doi.org/10.1145/2996890.3007867}

%Conference
%\conferenceinfo{PLDI '13}{June 16--19, 2013, Seattle, WA, USA}

%\acmPrice{\$15.00}

%
% --- Author Metadata here ---
%\conferenceinfo{WOODSTOCK}{'97 El Paso, Texas USA}
%\CopyrightYear{2007} % Allows default copyright year (20XX) to be over-ridden - IF NEED BE.
%\crdata{0-12345-67-8/90/01}  % Allows default copyright data (0-89791-88-6/97/05) to be over-ridden - IF NEED BE.
% --- End of Author Metadata ---

\title{The Effects of Relative Importance of User Constraints in \\Cloud of Things Resource Discovery: A Case Study}


\numberofauthors{2} %  in this sample file, there are a *total*
\author{
% 1st. author
\alignauthor 
Luiz H. Nunes\\
\affaddr{Federal Institute of S\~ao Paulo}\\
\affaddr{Mat\~ao-SP, Brazil}\\
\email{lhenriquenunes@ifsp.edu.br}
\alignauthor 
J\'ulio C. Estrella, Alexandre C. B. Delbem\\
\affaddr{University of S\~ao Paulo}\\
\affaddr{Institute of Mathematics and Computer Science}\\
\affaddr{S\~ao Carlos-SP, Brazil}
\email{$\{$jcezar, acbd$\}$@icmc.usp.br}\\
\and
\alignauthor 
Charith Perera\\
\affaddr{The Open University}\\
\affaddr{Walton Hall, Milton Keynes, MK7 6AA - UK}\\
\email{charith.perera@ieee.org}\\
\alignauthor 
Stephan Reiff-Marganiec\\
\affaddr{University of Leicester}\\
\affaddr{University Road, Leicester, LE1 7RH - UK}\\
\email{srm13@le.ac.uk }\\
}
% \alignauthor
% Luiz H. Nunes, J\'ulio C. Estrella, Charith Perera,\\ Stephan Reiff-Marganiec, Alexandre N. Delbem\\
% \affaddr{University of S\~ao Paulo,
% Institute of Mathematics and Computer Science,
% S\~ao Carlos-SP, Brazil} \\
% \email{$\{$lhnunes, jcezar, acbd$\}$@icmc.usp.br}\\
%       \affaddr{Instituto Federal de S\~ao Paulo , Mat\~ao-SP}\\
%       \email{lhenriquenunes@ifsp.edu.br}\\
%       \affaddr{The Open University, Walton Hall, Milton Keynes, MK7 6AA - UK}\\
%       \email{charith.perera@ieee.org}\\
%       \affaddr{University of Leicester,
% University Road, Leicester, LE1 7RH - UK}\\
%       \email{srm13@le.ac.uk }
% There's nothing stopping you putting the seventh, eighth, etc.
% author on the opening page (as the 'third row') but we ask,
% for aesthetic reasons that you place these 'additional authors'
% in the \additional authors block, viz.
% Just remember to make sure that the TOTAL number of authors
% is the number that will appear on the first page PLUS the
% number that will appear in the \additionalauthors section.

\maketitle
\begin{abstract}
Over the last few years, the number of smart objects connected to the Internet has grown exponentially in comparison to the number of services and applications. The integration between Cloud Computing and Internet of Things, named as Cloud of Things, plays a key role in managing the connected things, their data and services. One of the main challenges in Cloud of Things is the resource discovery of the smart objects and their reuse in different contexts. Most of the existent work uses some kind of multi-criteria decision analysis algorithm to perform the resource discovery, but do not evaluate the impact that the user constraints has in the final solution. In this paper, we analyse the behaviour of the SAW, TOPSIS and VIKOR multi-objective decision analyses algorithms and the impact of user constraints on them. We evaluated the quality of the proposed solutions using the \textit{Pareto-optimality} concept.
\end{abstract}



%
%  Use this command to print the description
%
\printccsdesc

% We no longer use \terms command


\section{Introduction}
%The Internet of Things (IoT) is a new revolution of the Internet, which considers the pervasive presence of objects capable of interact and cooperate with each other to achieve common goals using a common network in a specific environment. One of the most accepted definition is \textit{“The Internet of Things allows people and things to be connected Anytime, Anyplace, with Anything and Anyone, ideally using Any path/network and Any service.”} \cite{Vermesan:2013}, \cite{Perera:2014}. 

%The number of services and applications that apply the IoT concepts has been increasing in several areas such as environmental monitoring, healthcare aid and assisted car driving. - 

%On the other hand, the rapidly spread in the number and type of devices makes it difficult for IoT stakeholders to use the gathered data as generally they are collected for specific purposes. Several IoT solutions such as GSN and OpenIoT aims to provide the reuse of data to third parts.

Nowadays, the number of smart objects connected to the Internet is growing exponentially proportionally to the number of services and applications for them. According to the Gartner Report, there is about 6.4 billion of connected things moving a market around \$235 billion just with end-users services in 2016 \cite{Gartner:2015}. The integration between Cloud Computing and Internet of Things (IoT) named as Cloud of Things (CoT) plays a key role to manage the connected things, their data and the provided services \cite{Nunes:2016}.

%The recently increasing of smart objects connected in the Internet allowed to become the Internet of things in a reality in our lives. According to the Gartner Report, there is about 6.4 billion of connected things moving a market around \$235 billion just with end-users services. The integration of Cloud Computing and Internet of Things named as Cloud of Things is extremely important to handle with the connected things, their data and the provided services \cite{Nunes:2016}.

Jayaraman et al. \cite{Jayaraman:2016} defines the Cloud of Things paradigm ``\textit{where smart objects are fully connected to the network and integrated with the cloud(s) for data storage, processing, analytics and visualization}".
The number of services and applications using the CoT concepts has been increasing in several areas such as environmental monitoring, healthcare aid and assisted car driving. On the other hand, the rapidly spread in the number and type of devices makes it difficult for IoT stakeholders to use the gathered data, as generally they are used for specific purposes \cite{Sowe:2014}. 

Solutions such as GSN, OpenIoT and Xively aims to support the CoT vision enabling access, process and analyses of smart objects and their data by using a set of keywords or semantic inference. However, due to their dynamic nature and original goal, the data of a smart object could not be suitable to accomplish the requirements of a user different of its owner. 

%Context-aware computing allows us to store context1 information linked to sensor data so the interpretation can be done easily and more meaningfully. In addition, understanding context makes it easier to perform machine to machine communication as it is a core element in the IoT vision.

%Context-aware computing allows us to store context1 information linked to sensor data so the interpretation can be done easily and more meaningfully. In addition, understanding context makes it easier to perform machine to machine communication as it is a core element in the IoT vision.

%Context-aware computing concepts has been widely used to support sensor search and selection. It is capable of saving context information that characterise the IoT device state and link them to their data to make easier to reuse them according to user requirements (Charith Survey). Multiple-criteria decision analysis

The resource discovery process is a key challenge in the Cloud of Things context, which must to perform the smart objects search and selection regarding the constraints imposed by different users. In this sense, several research propose to use context-aware computing and multiple-criteria decision analysis (MCDA) to support the resource discovery process.

%The efficiently search and selection of set of proper devices respecting dynamic requirements of an user is an open challenge in IoT research. The Context-aware computing and multiple-criteria decision analysis concepts has been combined to support sensor search and selection. 

Context-aware computing refers to use stored context information to characterize a smart object and link them to their data \cite{Perera:2014}. While multiple-criteria decision analysis algorithms aims to propose the best set of smart objects according to the user objectives, constraints and their relative importance. The user constraints refer to the criteria imposed for the sensor discovery and the relative importance relate to the given weight of each criteria during the process.

Although several papers such as \cite{Nunes:2016, Gao:2014, Perera:2014a, Khodadadi:2015} use some kind of MCDA to perform the resource discovery process, they are not concerned about the quality of the proposed solution set. Moreover, they do not evaluate the impact of the relative importance of user constraints in the final set of smart objects. 

Thus, in this paper we present an evaluation of the impact of the user constraints and their relative importance to select a set of smart object. In particular, we investigate the behavior of the Simple Additive Weight method (SAW), the Technique for the Order of Prioritisation by Similarity to Ideal Solution (TOPSIS) and VIseKriterijumska Optimizacija I Kompromisno Resenje (VIKOR). To perform the experiments, the methodology presented in \cite{Nunes:2016b} is used to evaluate the algorithms and the impact of the user constraints in the final set.

The paper is organized as follows:
Section \ref{sec:RelatedWork} presents a literature review of resource discovery for CoT.
Section \ref{sec:MCDM} describes the analysed Multiple-criteria decision-making algorithms.
Section \ref{sec:PerformanceEvaluation} describes the methodology and configurations used to perform the experiments.
The results are then discussed in Section \ref{sec:Results}. Finally, the conclusions and directions for future work are presented in Section \ref{sec:Conclusion}.

\section{Related Work} \label{sec:RelatedWork} 
%Nos últimos anos diversas pesquisas vem sendo conduzidas nas diferentes áreas de IoT. Nesta Seção vamos apresentar trabalhos referentes aos anos de 2014 e 2015 que possibilitam ou realizam a busca e a Selection of dispositivos neste contexto.
Nowadays there are several approaches that enable the smart objects management. Perera et al. \cite{Perera:2014a} and R\"{o}mer et al. \cite{Romer:2010} present surveys that describes several architectures, techniques, methods, models, features, systems, applications, and middleware solutions related to the CoT context. In this section, firstly we present some architectures that enable the resource discovery of smart objects and next some works related to sensor discovery techniques.

%Antonic et al.~\cite{Antonic:2014} proposes an environment for colaborative mobile applications based on a middleware called CloUd-based PUblish/Subscribe, which uses a cloud environment to share context data with low energy consumption. Besides, this environment is capable to handle with the location and capacity of the mobile devices. It is able to gather data with regards to the desired QoS levels. A Broker application is deployed into the mobile devices to orquestrate the comunication between these devices and the cloud to save the energy resources.

Bovet and Hennebert~\cite{Bovet:2014} proposes a P2P architecture for sensor discovery aiming robustness, reliability and efficiency in energetic terms. The authors present an ontology to describe the properties, functionalities and how to access to the subscribed devices. The SPARQL language is used to look for specific devices into the ontologies, which are stored in a distributed manner over the nodes of the architecture.

Kamilaris et al.~\cite{Kamilaris:2014} use domain name server as a scalable metadata repository to support the entity discovery using their location. The authors proposes the creation of a new domain, such as \textit{.env}, which represent the entities of the real world. Thus, when a smart object becomes available it must register their characteristics and services into the DNS repository.

Kiljander et al.~\cite{Kiljander:2014} proposes an architecture aiming to provide smart objects interoperability. Ontologies are used to describe these devices which are accessible using SPARQL language and semantic agents. It uses unique identifiers named \textit{ucodes} to access and identify the devices of an specific network. The \textit{ucodes} are stored inside distributed brokers, which are organized according to their location, owners or data.

%Liu et al.~\cite{Mingliu:2014} presents an information architecture based on service for hardware constraint devices, which uses the cloud to process and store the gathered data. This architecture is composed by four layers. The perception layer aims to sensor and act in the environments; The network layer handles with the sensor in a local level and share the with the external environment. The service layer has the service providers and repositories that are used to access the information. The process layer analyse the requests, check the permissions and handles with the gathered data.

%Mrissa et al.~\cite{Mrissa:2014} proposes an avatar based architecture to integrate heterogeneous devices through semantics resources exposing them as a service. The avatars extends an object into the Web using an ontology that allow to interact with others avatars and external resources  and deploy their own application class.. This architecture is composed by a framework that uses a set of handlers that interact with each other. 

Diaz-Montes et al.~\cite{Diaz:2015} present the CometCloud tool to provide infrastructure and programming support to develop workflows to integrate with federated resources. CometCloud is a three-layer architecture composed by: infrastructure layer, autonomous management layer and interface layer. The infrastructure layer allows the information exchange with the distributed resources. The interface layer enables the information exchange between the user and the CometCloud core. Finally, the autonomous management layer compose the workflow according to available applications and their policies regarding the established SLA.

Carlson and Schrader~\cite{Carlson:2014} presents a search engine named Ambient Ocean to discovery and select sensors using context information. This search engine uses a local stored metadata to define the device context and perform the search in a more efficiently way. The search engine uses similarity multi-task models based on the Weighted Slope One algorithm. In scenarios that is hard to model the devices features, the Ambient Ocean applies collaborative filters techniques to compute the similarity between users or sensors using previous information.

Kothari et al.~\cite{Kothari:2014} shows an architecture named DQS-Cloud to optimize the sensor search, autonomous fault tolerance mechanism and avoid SLA violations. The search is based on keywords and in the QoS attributes desired by the users. The DQS-Cloud aims to minimize the communication overhead reusing data flows with similar QoS levels. The results shows that the DQS-Cloud was capable to minimize the bandwidth and processing rate in the providers.

%mediador
Gao et al.~\cite{Gao:2014} proposes the \textit{Automated Complex Event Implementation System} to integrate dataflows at runtime. The sensors and their flows are described according to the SSN ontology and are stored in a repository with their QoS and QoI attributes. %The system acts like a middleware between the sensor dataflow and an application. 
It is able to search and select the registered sensors with regards to the specified QoS and QoI levels using the Simple-Additive-Weighting algorithm. 

Perera et al. \cite{Perera:2014a} present the CASSARAM framework to perform the sensor search and selection regarding user context properties. It uses the Semantic Sensor Network Ontology (SSN) to retrieve and model user context properties. CASSARAM users use semi-negotiable context properties, which  allow to define context properties values in a range. Thus, the proposed Relational-Expression based Filtering can be applied to ignore irrelevant sensors during the semantic querying. Also, the Comparative-Priority Based Heuristic Filtering is used to remove the sensors that are far from the ideal point prioritizing the TOP-K selection.

%Datta et al.~\cite{Datta:2015} proposes a framework to create user-centric application to be executed in smartphones and tablets. This framework can dynamically search smart objects using semantics definitions and provide their sensors as a service. %It uses semantics to define the object application domain and enable their reuse for other purposes.The sensor search and selection uses the type of the sensor and the target domain. %The query result is presented to the user.

%busca
Doukas and Antonelli~\cite{Doukas:2015} presents the COMPOSE to provide an end-to-end solution to develop applications and services for CoT. This solution operates in all layers of IoT architecture interacting with the users of the mobile application, performing the sensor search and selection and also deploy the application into the cloud. The sensor search and selection uses the iServe, which is a service warehouse that unify several features such as the service publisher, service analyse and service discovery using semantics. The iServe is able to deploy service and additional features to explore the service description, notation and analysed gathered data.

%busca
Khodadadi et al.~\cite{Khodadadi:2015} proposes a framework named Simurgh to define ``things'', people and their functional properties to make easier define services and compose workflows for IoT. The search and selection process uses syntax based algorithms in two phases. The first phase, look for entities that respect a specific set of criteria. On the other hand, the second phase uses the first phase result set to perform another search to choose the suitable devices for a specific problem. The framework was validate using a study case that illustrated the framework behaviour for a temperature sensor.

Nunes et al.~\cite{Nunes:2016} presents the ViSIoT middleware to perform the smart objects resource discovery. This work use the TOPSIS to select the sensors according to the user constraints. The ViSIoT performance analyses shows the capacity for setting up the environment in a timely manner. 

The discussion presented in this section show some architectures and alternatives to perform the resource discovery and selection according to the constraints imposed by the final user. However, these works do not evaluate the quality of the proposed solutions and the effects of the user constraints in the quality of the proposed smart objects set. In this sense, we have take the SAW and TOPSIS algorithms presented in this section  as a base algorithm for a case study about the efficiency of MCDA algorithms applied in the CoT context and analyse the influence of user constraints in the final solution.

%The majority of the related work presented in this section presented architectures to enable the integration of sensors and doesn't discuss how the sensor search and selection are realized. A small number of works have presented alternatives to perform the sensor and selection of sensors, but they didn't analysed the quality provided by these methods. In this paper we apply three popular MCDM algorithms into the IoT context and analyse the quality of the selection provided by then. Also, we assume that the context properties can have different priorities.

\section{Multi-Objective Optimization} \label{sec:MCDM}

Several problems in industry, computing, engineering and other areas uses multiple objectives optimization. In many cases, these objectives are defined in not comparable units and have some level of conflict between them. In other words, an objective can not be improved without deteriorate another objective \cite{Jaimes:2009}. In the sensor discovery process this scenario can be exemplified by an user which desires to choose a subset of smart objects but also wants to minimize the price and maximize the accuracy of the sensors in this subset.

In an optimization problem with one objective, the search space is always well defined. As more conflicting goals must be simultaneously optimized, it is extremely hard to establish a single optimal solution but rather a set of possibilities with equivalent quality. The optimal solution is a set of optimal trade-offs between conflicting goals \cite{Abraham:2005}.
An multi-objective optimization problem can be describe as
\begin{align}
\text{minimize} \{f_{1}(x), f_{2}(x), \cdots, f_{k}(x)\} \text{, where } x \in S,
\end{align}

wherein the number of objective functions $ k $ is greater than or equal to two in decision space, represented by $ R  ^ {n}$. 
The vector of objective functions is defined by:
\begin{align}
f(x) = (f_ {1} (x), f_ {2} (x), \cdots, f_ {k} (x)) ^ {T}
\end{align}
Vector decision $ x = (x_ {1}, x_ {2}, \cdots, x_ {n}) ^ {T} $ belong to feasible region nonzero $ S $, which is a subset of $ R ^ {n} $ \cite{Miettinen:2012}.

\subsection{Pareto Optimality}

The Pareto dominance relationships are used to compare different sets of solutions. The set of optimal solutions of problem is given the name of set of optimal solutions or Pareto non-dominated solutions \cite{Jaimes:2009}.

In a minimization problem, a solution $x^{T}$ is not dominated if there is no $x \in S$  such that $f_{i}(x) \leq f_{i}(x^{T})$ for each objective$_{i = 1, \cdots, k}$ e $f_{i}(x) < f_{i}(x^{T})$ for at least one of the analyzed objectives\cite{Abraham:2005}. The image of the set of optimal solutions is called Pareto frontier or Pareto curve. The shape of the Pareto front indicates the nature of trade-off between different objective functions \cite{Caramia:2008}.

Each objective can be minimized or minimized.  The solid curve represents the set of non-dominated solutions. It this figure, the optimal set of Pareto is always composed of the solutions that are concentrated in an specific vertex of a feasible region. Furthermore, in a continuous space of solutions, the optimal set of Pareto may be formed by two disjoint sets of solutions as represented by Figure(b). However, despite the existence of multiple Pareto optimal solutions in practice only one of these solutions must be used \cite{Deb:2001}.

\subsection{Multiple-criteria decision analysis} 

Multiple-criteria decision analysis algorithms are used for decision making in the presence of multiple and often conflicting goals. The MCDA algorithm are intended to assist the judgment of decision making through a set of goals and criteria, estimating their importance and establishing the contribution of each option regarding a set of criteria \cite{Dodgson:2009}. 

An MCDA problem can be described using an analysis  matrix $M \times N$, where the element $q_{ij}$ represents the performance of each option according to the decision criteria $c_{j}$ in non comparable units and scales, such as represented by Equation~\ref{eq:matriz}. An evaluation matrix is used to represent the relative performance of each $q'_{ij}$ using a  normalization function to compare the different criteria \cite{Tzeng:2011}.

\begin{align}
    Q = 
    \begin{blockarray}{cccccc} 
    &c_{1} & c_{2} & c_{3} &  & c_{n} \\
        \begin{block}{c[ccccc]}
          q_{1} & q_{11} & q_{12} & q_{13} & \hdots & q_{1n} \\    
          q_{2} & q_{21} & q_{22} & q_{23} & \hdots & q_{2n} \\
                & \vdots &\vdots  &\vdots & \vdots & \vdots \\
          q_{m} & q_{m1} & q_{m2} & q_{m3} & \hdots & q_{mn} \\
        \end{block}
    \end{blockarray}
    \label{eq:matriz}
 \end{align}

All MCDA algorithms  explicitly define its options and weights of each criterion, but differ in the way that they combine the input data. Although MCDA problems are found in different areas, they often share similar characteristics such as using multiple criteria always form a hierarchy , conflict between the criteria, hybrid nature, uncertainty and their solutions can not be conclusive \cite{Xu:2001}.

\subsubsection{SAW}

The \textit{Simple Additive Weight} algorithm is one of the most popular MCDA algorithms and is applied in several application domains such as supply chain management, personnel selection problems, project manager selection and facility location selection \cite{Abdullah:2014}, \cite{Ramezani:2012} . The SAW algorithm aims to get a weighted sum of the normalized criterion values of each alternative, where the greater value represents the preferred alternative \cite{Ramezani:2012}.

\subsubsection{TOPSIS}

    The \textit{Technique for Order of Preference by Similarity to Ideal Solution} is another popular MCDA algorithm applied in Supply Chain Management and Logistics, Design, Engineering and Manufacturing Systems,  Business and Marketing Management, Health, Safety and Environment Management, Human Resources Management, Energy Management, Chemical Engineering and Water Resources Management \cite{Behzadian:2012}. The TOPSIS algorithm aims to choose the options that are closest to the optimal solution and farthest from the negative optimal solution \cite{Tzeng:2011}.
    
\subsubsection{VIKOR}
The \textit{VIseKriterijumska Optimizacija I Kompromisno Resenje} uses the concept of compromise programming and has been applied in several fields such as location selection, environmental policy and data envelopment analysis \cite{Huang:2009}. The VIKOR algorithm aims to find the options that are closest to the optimal solution, and also evaluate their individual and group impact \cite{Tsai:2011}.

\section{Evaluation Methodology}\label{sec:PerformanceEvaluation}
This Section presents the research methodology used in the experiments. We use the evaluation methodology proposed by Nunes et al. \cite{Nunes:2016b} to compare resource discovery algorithms from a quality of search perspective. In this Section we assume the criteria and user constraints have the same meaning and weights as well as  relative importance.

\begin{figure}[!htp]
\centering
\includegraphics[scale=0.31]{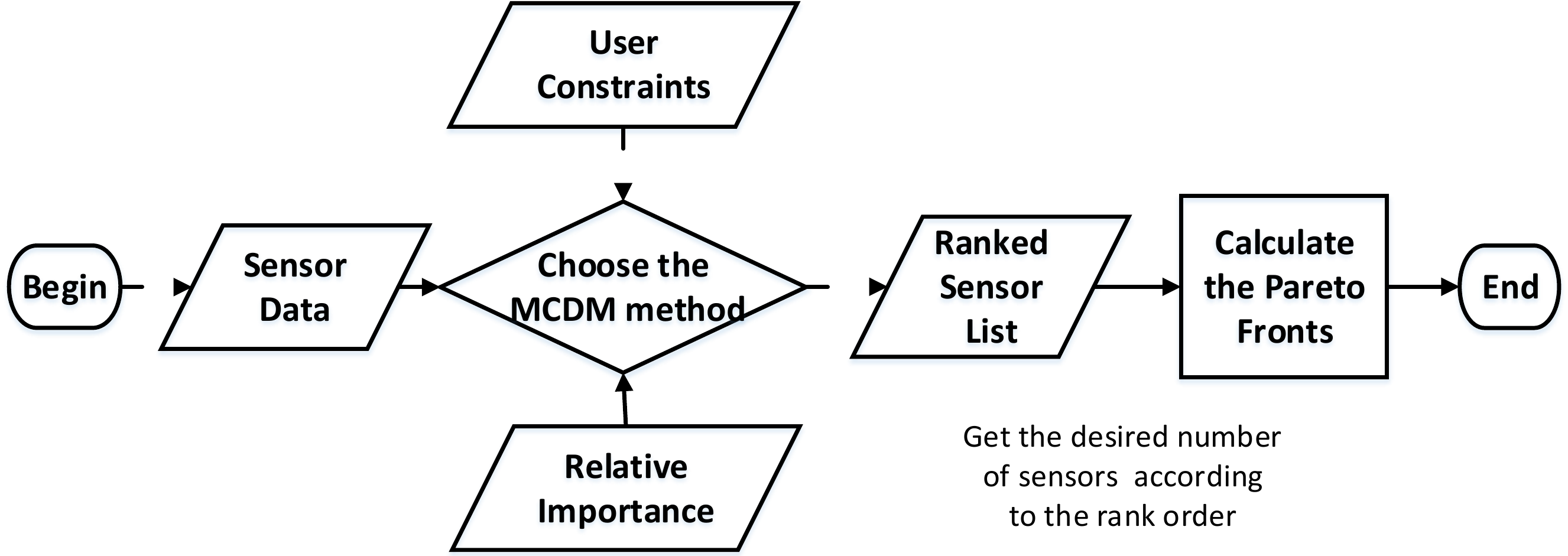}
\caption{Evaluation Workflow. Adapted from Nunes et al. \cite{Nunes:2016b}}
\label{pic:workflow}
\end{figure}

Figure~\ref{pic:workflow} shows the workflow used in our experiments. The sensor data, the user constraints and their relative importance are used as input for a MCDA algorithm that will output a ranked sensor list. Next, the ranked sensor list is used as input for the Pareto Optimal Solutions Check, which define the number of optimal solutions in each Pareto front. 

 The metric used to evaluate the MCDA algorithms is the the Overall non-dominated vector generation ratio (ONVGR)~\cite{Collette:2004} which shows the number of optimal solutions in the Pareto front as a proportion of the number of solutions proposed by the MCDA methods in each front. The closer to one the ONVGR value is, the better is the solution proposed in that front.

The experiment environment is composed by only one physical machine, which hosts the application with multi-criteria decision algorithms. Table~\ref{tab:environment} describes the hardware used to execute the algorithms.

\begin{table}[htbp]
  \centering
  \caption{Physical Environment}
  \begin{adjustbox}{max width=0.47\textwidth}
    \begin{tabular}{cc}
    \hline
    \textbf{Hardware/Software} & \textbf{}{\textbf{Specification}} \\
    \hline
    \multicolumn{1}{c}{Processador} & \multicolumn{1}{c}{AMD Processor Vishera 4.2 Ghz} \\
    \multicolumn{1}{c}{Memory} & \multicolumn{1}{c}{32 GB RAM DDR3 Corsair Vegeance} \\
    \multicolumn{1}{c}{Hard Disk} & \multicolumn{1}{c}{HD 2TB Seagate Sata III 7200RPM} \\
    \multicolumn{1}{c}{Operating System} & \multicolumn{1}{c}{Linux Ubuntu Server 14.04 64 Bits LTS} \\
    \multicolumn{1}{c}{Java} & \multicolumn{1}{c}{JDK 1.7} \\
    \multicolumn{1}{c}{Database} & \multicolumn{1}{c}{MongoDB 3.0} \\
    \hline
    \end{tabular}%
    \end{adjustbox}
  \label{tab:environment}%
\end{table}%

The experimental methodology was based on four factors: i) the number of sensors descriptions, ii) the MCDA algorithm, iii) the number of selected sensors and iv) the number of criteria. Table \ref{tab:exp} shows the used experimental factors and levels, where the combination of the levels of each factor gives a total of 12 experiments. Each experiment was replicated one hundred times, where the criteria weights was randomly defined at execution time.

\begin{table}[htbp]
  \centering
  \caption{Factors and levels used in the experiment}
  \begin{adjustbox}{max width=0.47\textwidth}
    \begin{tabular}{rc}
    \hline
    \multicolumn{1}{c}{\textbf{Factor}} & \textbf{Level} \\
    \hline
    Number of Sensors Descriptions & 100,000 \\
    MCDA Method & SAW, TOPSIS and VIKOR \\
    Number of Selected Sensors & 1,000 and 10,000 \\
    Number of User Constraint & 2 and 6 \\
    \hline
    \end{tabular}%
    \end{adjustbox}
  \label{tab:exp}%
\end{table}%

The sensor descriptions used as algorithm input was synthetically generated. The sensor capabilities and measurements (e.g. frequency and power consumption) are based on the 4027A Series from Bird Technologies\footnote[1]{Bird Technologies  -http://www.birdrf.com/}. The context data related to each sensor are retrieved from OpenWeatherMap\footnote[2]{OpenWeatherMap - http://openweathermap.org/} and their current properties values used in this experiment (e.g. battery, price, drift and response time) are assumed to be retrieved by software systems that manage such data and are available to be used.

The user constraints and objectives functions used to maximize (max($c_{j}$)) or minimize (min($c_{j}$)) follow this order: max(battery), min(price), min(drift), max(frequency),    min(\\energy consumption), min(response time).

\section{Results}\label{sec:Results}

\begin{figure*}[ht]
\centering
\subfigure[Selection of 1\%]{\includegraphics[scale=0.30]{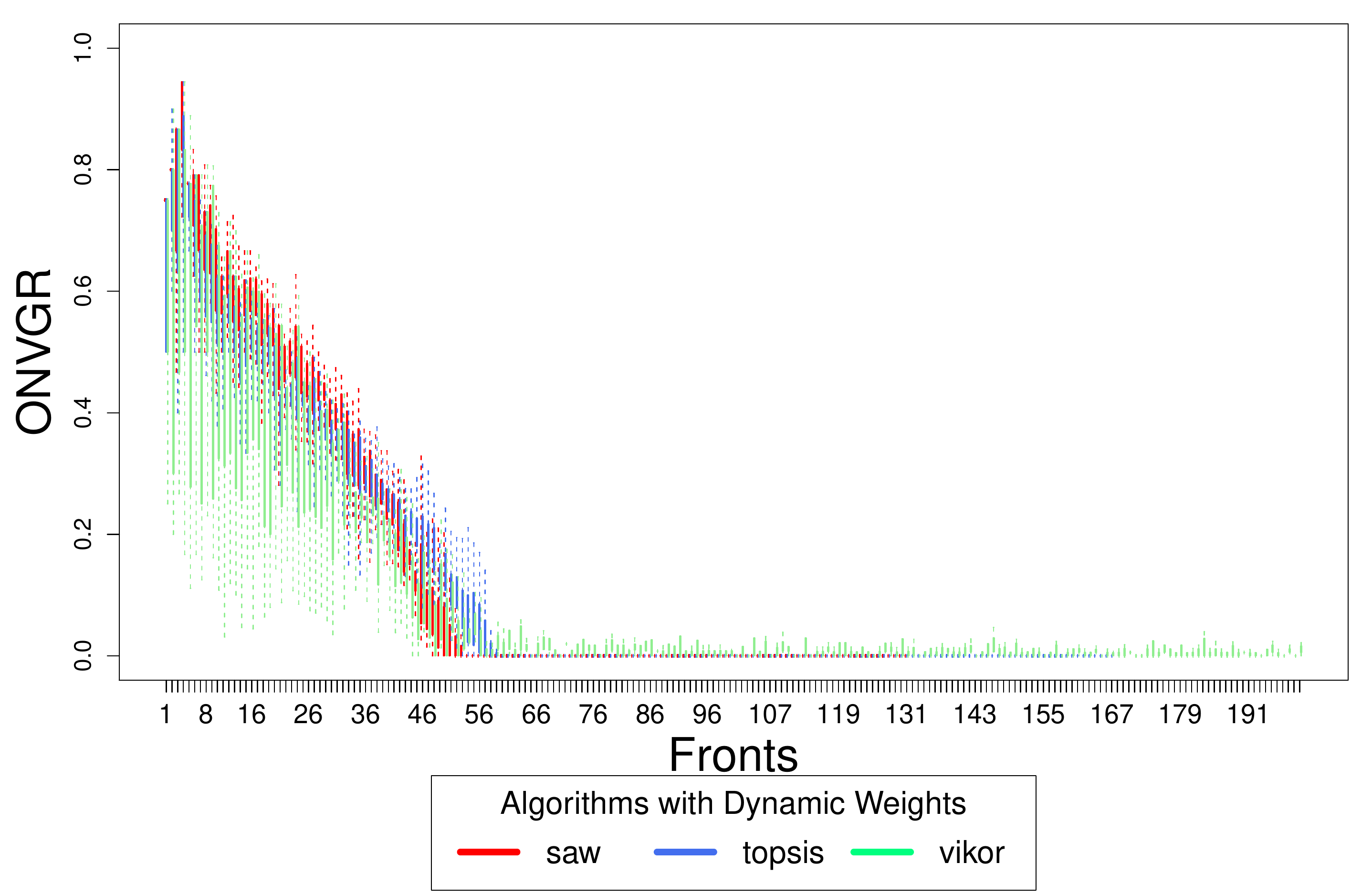}}
\subfigure[Selection of 10\%]{\includegraphics[scale=0.30]{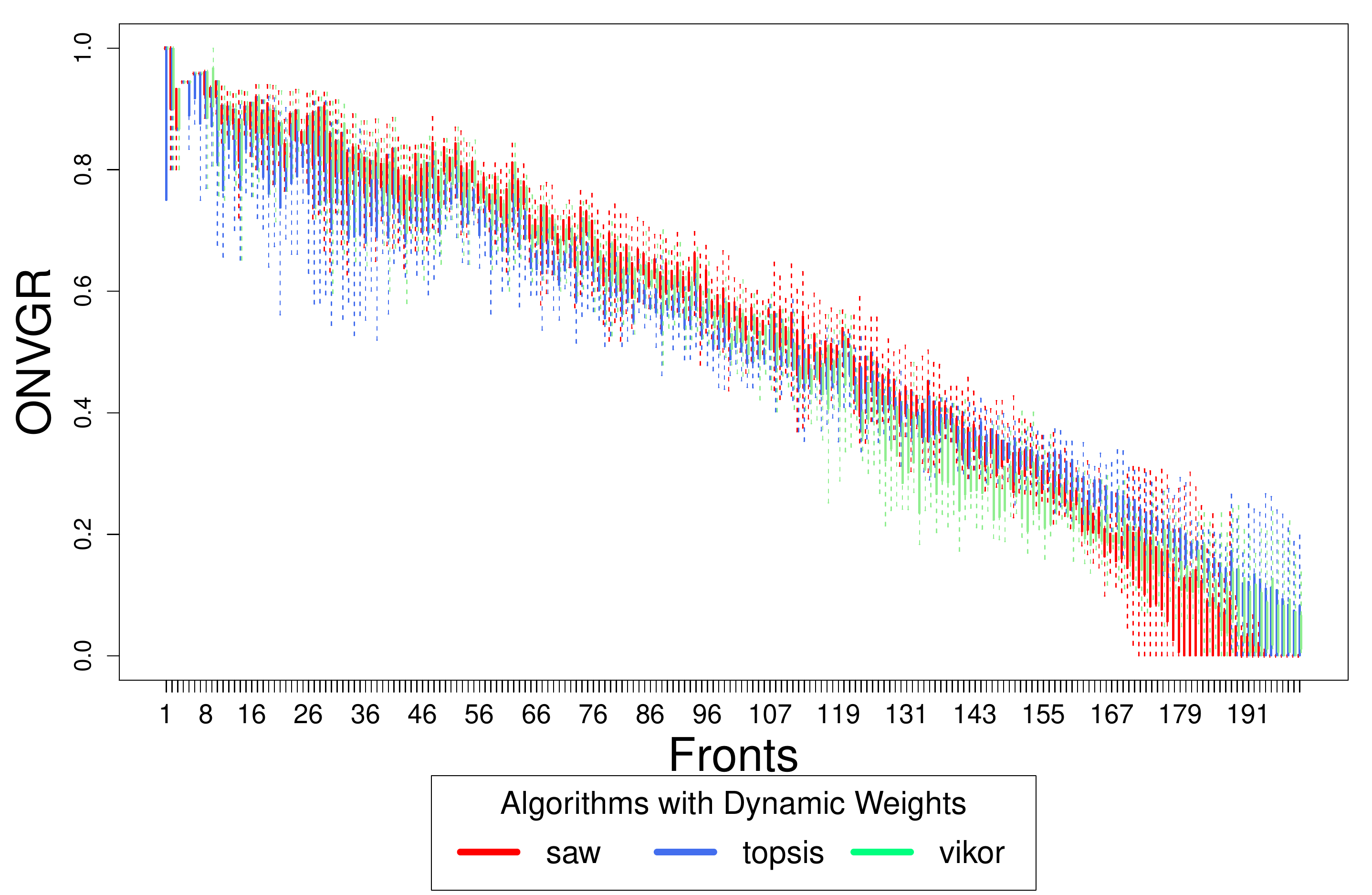}}
\caption{Variation of ONVGR value for two user constraints}
\label{boxplot:2}
\end{figure*}

Figure~\ref{boxplot:2} presents the boxplot representation of the ONVGR to select 1,000 (Figure~\ref{boxplot:2}.a) and 10,000 (Figure~\ref{boxplot:2}.b) smart objects considering two user constraints. In this figure we have to suppress the outliers and limit the number of fronts to two hundred to allow the graphic view. We observe that are a high number of fronts due to the low number of user constraints conflicts, which impacts in the number of available solutions in each front. 

\begin{figure*}[ht]
\centering
\subfigure[Selection of 1\%]{\includegraphics[scale=0.30]{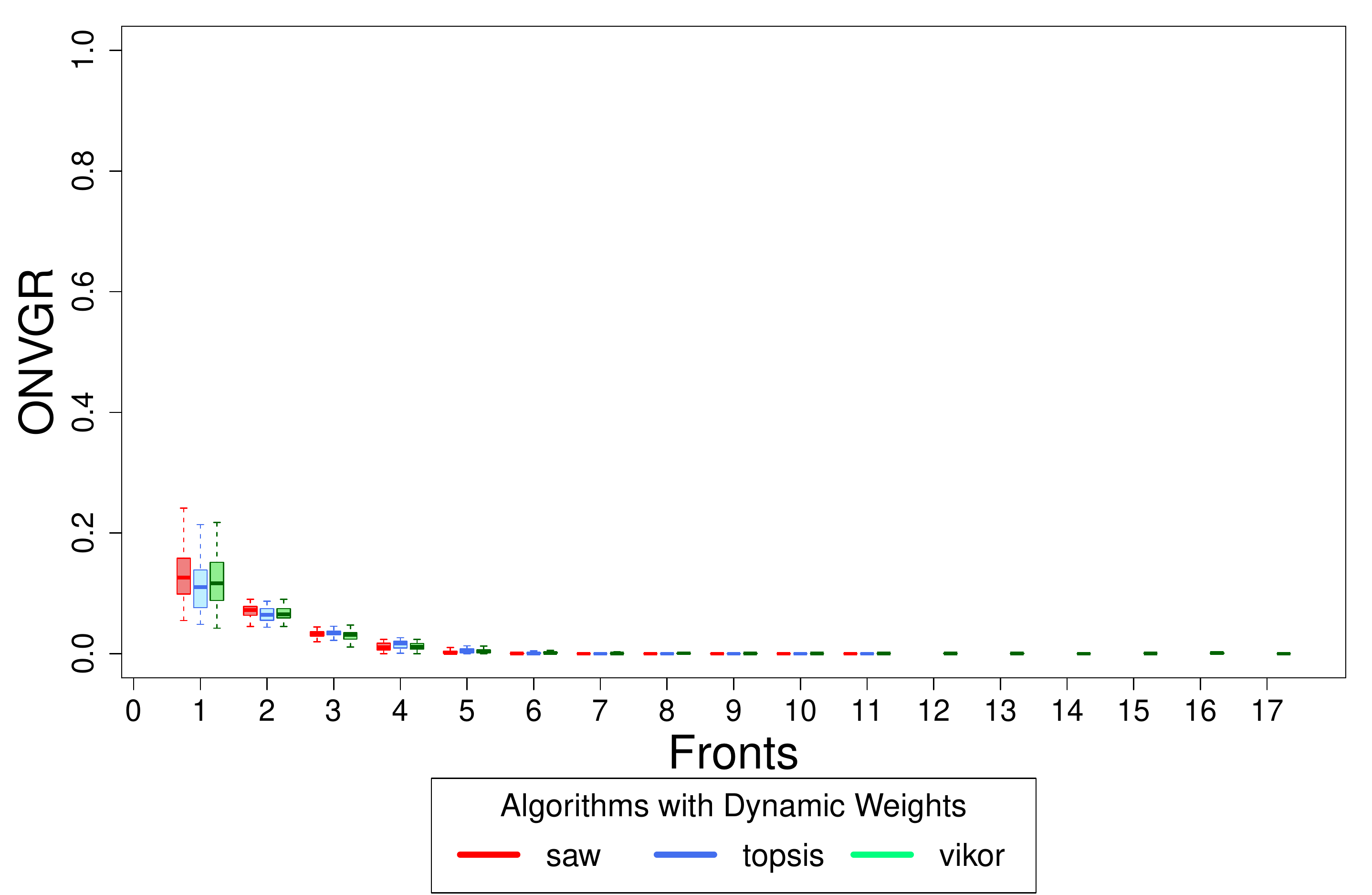}}
\subfigure[Selection of 10\%]{\includegraphics[scale=0.30]{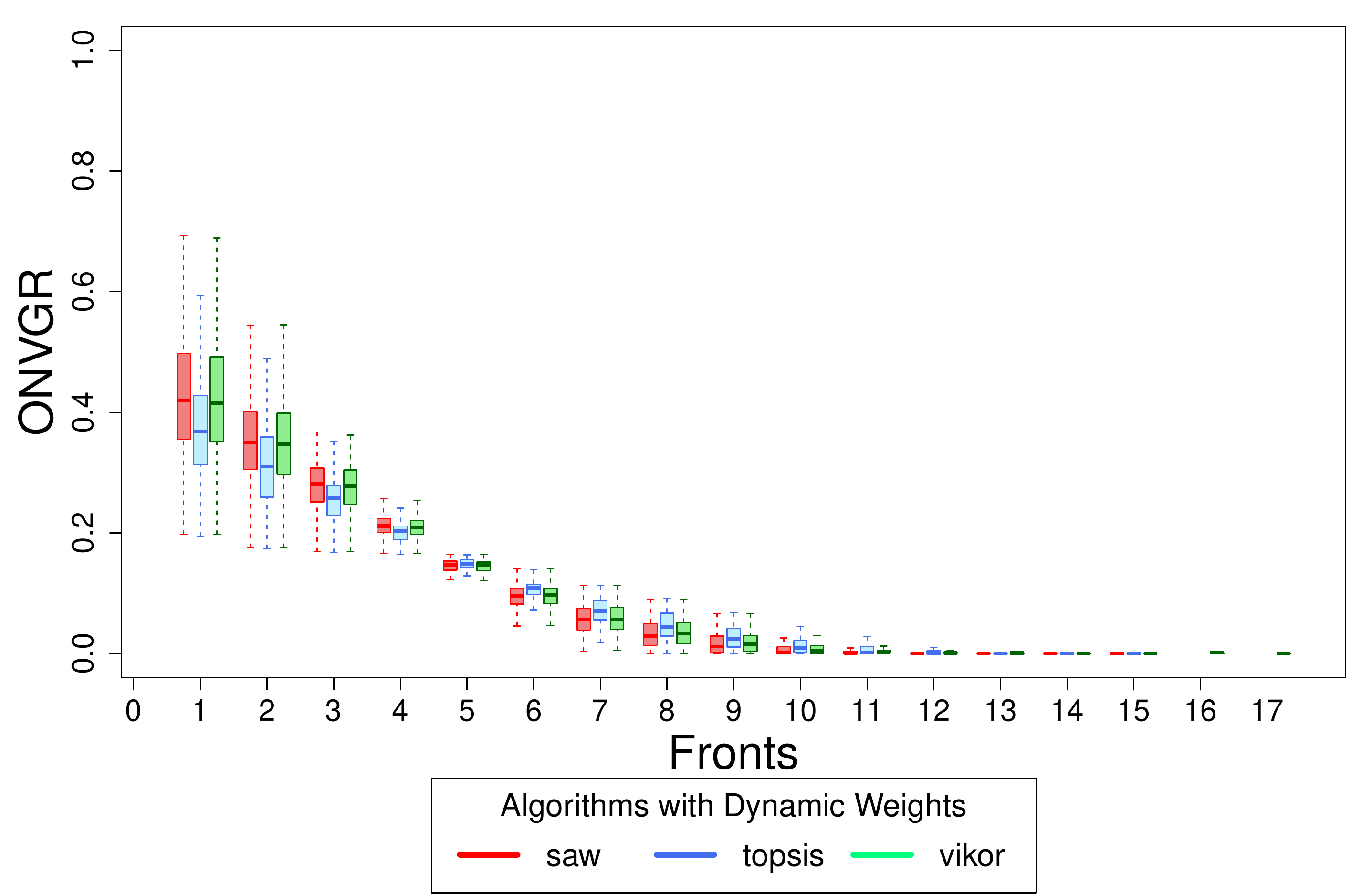}}
\caption{Variation of ONVGR value for six user constraints}
\label{boxplot:6}
\end{figure*}
The number of solutions available in each front increases proportionally to front index which decrease the ONVGR value because less objects are selected in these fronts.
Also, the ONVGR value has a low variation when 10\% of the sensors are selected rather than 1\%, because more sensors are selected which increases the chances of select an optimal sensor independently of the user constraints. 
The  mean behaviour of the algorithms shows that they could not find all the optimal smart objects, in which in the best scenario an ONVGR value closer to 0.8 and 0.9  are got when 1\% and 10\% of the smart objects are selected. Regarding the MCDA algorithms, we can observe the boxplot overlap each other, thus they present an equivalent behaviour when user constraints are used.

%We observe the ONVGR value is high in the first front and get low as the front index increase because it handle with the minimal possible criteria conflict in this scenario.
Figure~\ref{boxplot:6} presents the graphic representation of the ONVGR to select 1,000 (Figure~\ref{boxplot:6}.a) and 10,000 (Figure~\ref{boxplot:6}.b) smart objects considering six user constraints. In this Figure we observe that are less fronts than the solution presented in Figure~\ref{boxplot:6} because there are more conflicts between the user constraints and consequently a high number of optimal solutions in each front. Thus, the ONVGR value is lower than the two properties scenario as the algorithms are not able to find all the solutions in the first fronts. 
It is important to highlight when 1\% and 10\% of the smart objects are selected  a mean ONVGR value lower than 0.2 and around 0.4 are gotten. In this sense, the quality of the proposed solution when six user constraints are considered is worst than the proposed solution with two user constraints. About the MCDA algorithms, the three algorithms present practically the same behaviour with a slightly difference for VIKOR algorithm which uses two more fronts than SAW and VIKOR with a low number of solutions for each one.

In summary, the results show that the use of relative importance in user constraints does not necessarily improve the quality of the solution offered by a MCDA algorithm regarding the Pareto dominance relationships. The change of relative importance in user constraints has a higher impact when less constraints are used because there are less optimal solutions in each front. As expected, as more user constraints are used worst is the proposed solution. Finally, in the analysed scenarios when the relative importance of user constraints are changed there is no statistical difference between the solutions proposed by the MCDA algorithms.

%991714350 lemao
\vspace{1cm}
\section{Conclusion}\label{sec:Conclusion}
Efficient resource discovery of smart objects by adhering to dynamic requirements of an user is an open challenge in CoT environments. The integration of context-aware computing and multi-objective optimization has been widely used to support the sensor search and selection and to find the best trade-off between the available solution and the imposed constraints. In this paper, we have used an existent methodology presented in Nunes et al. \cite{Nunes:2016b} to evaluate MCDA algorithms and the impact of the relative importance of user constraints in the quality of the proposed smart object results set. The gathered results show that the use of relative importance in user constraints does not necessarily improve the quality of the solution offered by a MCDA algorithm and has a higher impact when less user constraints are used. Further, the higher number of user constraints decreases the quality of the proposed solution due to conflicts between user constraints. For future work, we will analyze other characteristics of their solutions such as convergence and distribution regarding the Pareto dominance relationships.

\bibliographystyle{abbrv}
\bibliography{sigproc}  % sigproc.bib is the name of the Bibliography in this case
% ACM needs 'a single self-contained file'!
%\balancecolumns
\end{document}